\documentclass{article}
\usepackage{amssymb}
\usepackage{graphicx}
\usepackage{amsmath}
\usepackage{adjustbox}
\usepackage[table]{xcolor}
\usepackage{comment}
\usepackage[pagebackref,breaklinks,colorlinks,allcolors=cvprblue]{hyperref}
\usepackage{wrapfig}
\usepackage{booktabs}
\usepackage{enumitem}

\usepackage[final]{corl_2025} 

\definecolor{ogreen}{HTML}{2E7D32}

\hypersetup{
    citecolor=ogreen
}

\title{
HyperTASR: Hypernetwork-Driven Task-Aware Scene Representations for Robust Manipulation
}

\author{
  Li Sun$^{*}$, \ \ Jiefeng Wu$^{*}$, \ \ Feng Chen, \ \ Ruizhe Liu, \ \ Yanchao Yang\\
  \\
  Institute of Data Science \& Department of Electrical and Electronic Engineering\\
  \smallskip
  The University of Hong Kong\\
  \texttt{\{sunlids, jiefengwu, cf24, zrllrz360\}@connect.hku.hk, yanchaoy@hku.hk}\\
}

\begin{document}
\maketitle
\let\thefootnote\relax\footnote{$*$: equal contribution.}

\vspace{-5mm}
\begin{abstract}
Effective policy learning for robotic manipulation requires scene representations that selectively capture task-relevant environmental features. Current approaches typically employ task-agnostic representation extraction, failing to emulate the dynamic perceptual adaptation observed in human cognition. We present HyperTASR, a hypernetwork-driven framework that modulates scene representations based on both task objectives and the execution phase. Our architecture dynamically generates representation transformation parameters conditioned on task specifications and progression state, enabling representations to evolve contextually throughout task execution. This approach maintains architectural compatibility with existing policy learning frameworks while fundamentally reconfiguring how visual features are processed. Unlike methods that simply concatenate or fuse task embeddings with task-agnostic representations, HyperTASR establishes computational separation between task-contextual and state-dependent processing paths, enhancing learning efficiency and representational quality. Comprehensive evaluations in both simulation and real-world environments demonstrate substantial performance improvements across different representation paradigms. Through ablation studies and attention visualization, we confirm that our approach selectively prioritizes task-relevant scene information, closely mirroring human adaptive perception during manipulation tasks. The project website is at \href{https://lisunphil.github.io/HyperTASR_projectpage/}{lisunphil.github.io/HyperTASR\_projectpage}.
\end{abstract}

\section{Introduction}
\label{sec:intro}

Embodied AI
has made significant advances in recent years~\cite{pfeifer1998representation,duan2022survey,liu2024aligning,xu2024survey}, 
driven by the mission of creating intelligent agents
that can interact with physical environments
with both effectiveness and robustness. 
These capabilities are essential for numerous real-world applications, 
necessitating the development of
generalizable policy learning frameworks that translate
perceptual observations into precise motor commands~\cite{mu2024embodiedgpt}. 
A {\it typical policy learning} pipeline 
comprises a representation extraction module 
that transforms raw observations into structured scene representations 
and a policy module 
that maps these representations to actions~\cite{xu2020learning,billard2019trends,li20223d}.

\begin{figure}[!t]
    \centering
    \includegraphics[width=1\linewidth]{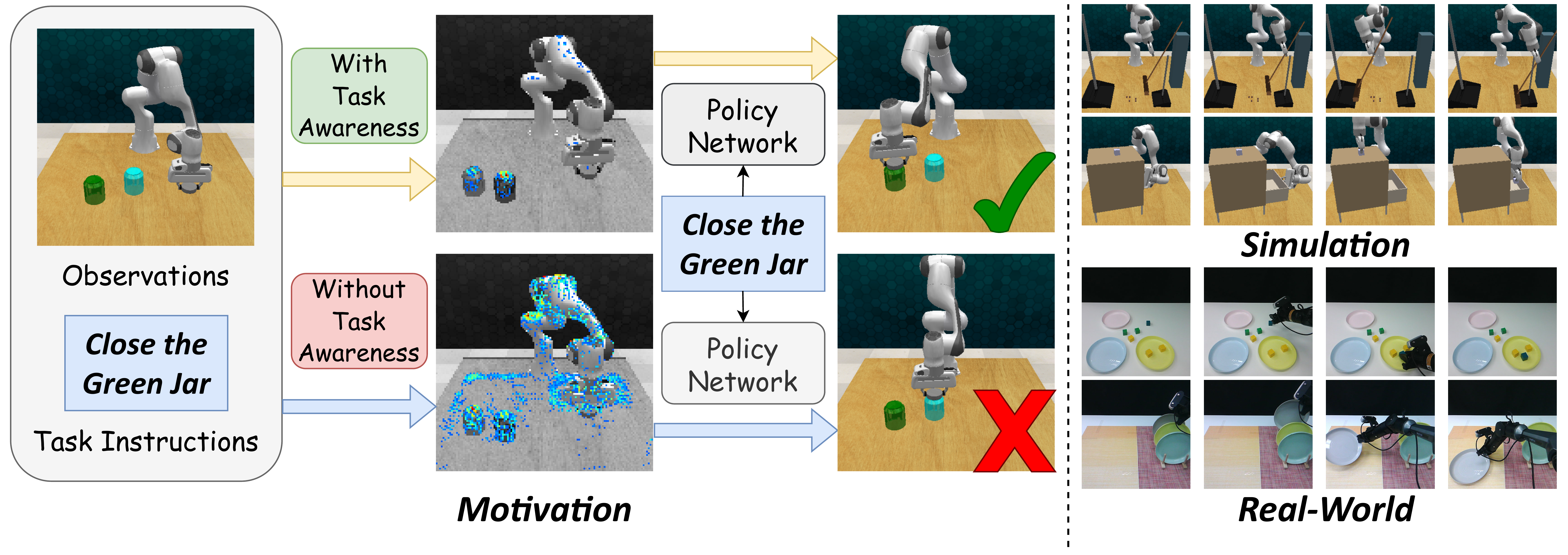}
    \vspace{-7mm}
    \caption{
    Task-aware representations enable selective attention to task-relevant scene elements, enhancing manipulation performance.
    {\it Top-left:} Our proposed HyperTASR pipeline incorporates task-aware scene representation extraction that dynamically modulates feature processing based on both task objectives and execution phase.
    {\it Bottom-left:} Conventional pipelines employ fixed, task-agnostic representation extractors that process visual information uniformly across all tasks, limiting representational flexibility.
    {\it Right:} Visualization of manipulation tasks in sim \& real.}
    \vspace{-3mm}
    \label{fig:pips-compare}
\end{figure} 

To enable flexible interactions across diverse scenarios, 
modern policy architectures incorporate task conditioning, 
enabling multi-task learning capabilities
that facilitate the sharing of transferable skills -- a
critical step toward general-purpose embodied intelligence. 
When trained end-to-end on demonstration data, 
these pipelines have demonstrated impressive performance across various manipulation tasks and reasonable robustness in novel scenarios.

However,
current policy learning approaches 
typically implement scene representation extraction 
as a {\it task-agnostic} process,
which is decoupled from action prediction (Fig.~\ref{fig:pips-compare}). 
This separation omits established findings in human cognition, 
where extensive research demonstrates that visual processing adaptively reconfigures based on task objectives and execution context. 
Studies by Hayhoe \cite{hayhoe2003visual,hayhoe2017vision}, Rothkopf \cite{rothkopf2007task}, and Foulsham \cite{foulsham2011and} reveal how visual representations dynamically adjust to task demands. 
This adaptation aligns with Gibson's~\cite{gibson2014ecological} concept of affordances and the Theory of Event Coding (TEC)~\cite{hommel2001theory}, as well as neurological evidence on adaptive neural representations~\cite{jamone2016affordances}.

Motivated by these insights, we propose \textit{HyperTASR} -- a hypernetwork-driven task-aware scene representation framework 
that enables policy learning pipelines to selectively focus on task-relevant scene elements, thereby enhancing sample efficiency and generalization capabilities.
Consider the progressive nature of cup grasping: 
initially requiring coarse spatial awareness for localization, 
then transitioning to fine-grained geometric perception as the gripper approaches the handle. 
This dynamic modulation of representational focus facilitates precise and efficient scene interaction.

To {\it implement} this task-conditional scene representation, 
we introduce a modular transformation framework 
that adaptively reconfigures representations based on both task objectives and execution phase. 
{\it Specifically,} we employ a hypernetwork architecture that dynamically generates the parameters 
of a representation transformation network conditioned on task specifications and progression state. 
This approach establishes computational separation between task-contextual and state-dependent gradients \cite{sarafian2021recomposing}, 
significantly enhancing learning efficiency. 
The framework continuously modulates scene representations throughout task execution,
ensuring that the extracted features remain optimally aligned with current manipulation requirements.

We integrate HyperTASR with 
two representative state-of-the-art policy learning architectures: 
one employing train-from-scratch representations~\cite{Ze2023GNFactor} and another utilizing fixed pre-trained backbones~\cite{ke20243d}. 
In \textbf{simulation} experiments on RLBench~\cite{james2020rlbench}, 
our framework substantially elevates performance across both architectures. 
{\it Notably,} integration with GNFactor increases success rates by more than 27\%, while implementation with 3D Diffuser Actor achieves success rates exceeding 80\% for the first time in single-view configurations. In \textbf{real-world} experiments, HyperTASR enables effective multi-task manipulation even with limited demonstration data, {\it outperforming} baseline methods.
Through comparative analysis with ablated models and attention visualization, we demonstrate that our approach selectively prioritizes task-relevant scene information throughout execution.

In summary, our contributions are: 
\begin{itemize}[leftmargin=*,noitemsep]
    \vspace{-3mm}
    \item We propose HyperTASR, a novel framework for extracting task-aware scene representations that enables robotic agents to emulate human-like adaptive perception by focusing on the most task-relevant environmental features throughout execution.
    \item We introduce a hypernetwork-based representation transformation that dynamically generates adaptation parameters conditioned on both task specifications and progression state, maintaining architectural compatibility with existing policy learning frameworks.
    \item We demonstrate through comprehensive experiments in both simulation and real-world settings that HyperTASR significantly enhances performance across different representation paradigms, establishing new state-of-the-art results for single-view manipulation.
\end{itemize}

\section{Related Work}
\label{sec:relatedwork}

\paragraph{Scene Representation for Multi-task Robotic Manipulation.} 
Recent advancements in multi-task robotic manipulation have significantly improved task execution and generalization~\cite{shridhar2023perceiver,brohan2022rt,jang2022bc,shridhar2022cliport,goyal2023rvt,goyal2024rvt,james2022coarse}. 
The dominant paradigm extracts scene representations from sensory input for action mapping~\cite{kroemer2021review,nair2022r3m,zhang2020polarnet}. 
State-of-the-art approaches~\cite{shridhar2023perceiver,gervet2023act3d,lu2025manigaussian} leverage foundation models to inject semantic knowledge, with methods like~\cite{ke20243d} directly utilizing pretrained visual backbones and GNFactor~\cite{Ze2023GNFactor} incorporating feature distillation -- enhancing generalizability across environments~\cite{karamcheti2023language}.
However, a fundamental limitation persists: scene representations typically remain task-agnostic and static throughout execution. This contradicts findings in human cognition~\cite{hayhoe2003visual,rothkopf2007task}, where visual processing dynamically reconfigures based on task objectives and context. While Vision-Language-Action models~\cite{o2024open,kim2024openvla,team2024octo} incorporate task information through language embeddings, few integrate task context at the representation stage. RT-1~\cite{brohan2022rt} employs FiLM~\cite{perez2018film} to encode instructions alongside observations but lacks explicit representation learning.
Our approach fundamentally differs by conceptualizing representation adaptation as a dynamic transformation process that evolves throughout task execution -- a significant advance beyond methods that maintain static representations or implement simple conditioning without accounting for temporal context.

\paragraph{Hypernetworks in Robotic Learning.} 

Hypernetworks~\cite{ha2017hypernetworks} provide an efficient framework for implementing our task-aware scene representations. Developed for neural architecture search~\cite{brock2018smash,zhang2018graph}, continual learning~\cite{von2020continual,henning2021posterior}, generative modeling~\cite{ratzlaff2019hypergan,skorokhodov2021adversarial}, and reinforcement learning~\cite{huang2021continual,xian2021hyperdynamics,rezaei2023hypernetworks,beukman2024dynamics}, hypernetworks excel at generating specialized parameters conditioned on task-specific information~\cite{chauhan2024brief}. In robotic applications~\cite{beck2023hypernetworks,renhypogen}, they enable adaptation across diverse scenarios. We leverage hypernetworks for their advantages in enabling functional transformation of representation spaces rather than simple feature weighting -- particularly well-suited for realizing our core contribution of adaptive perceptual processing that continuously evolves throughout task execution.

\section{Method}
\label{sec:method}

\begin{figure}[!tb]
  \centering
   \includegraphics[width=\linewidth]{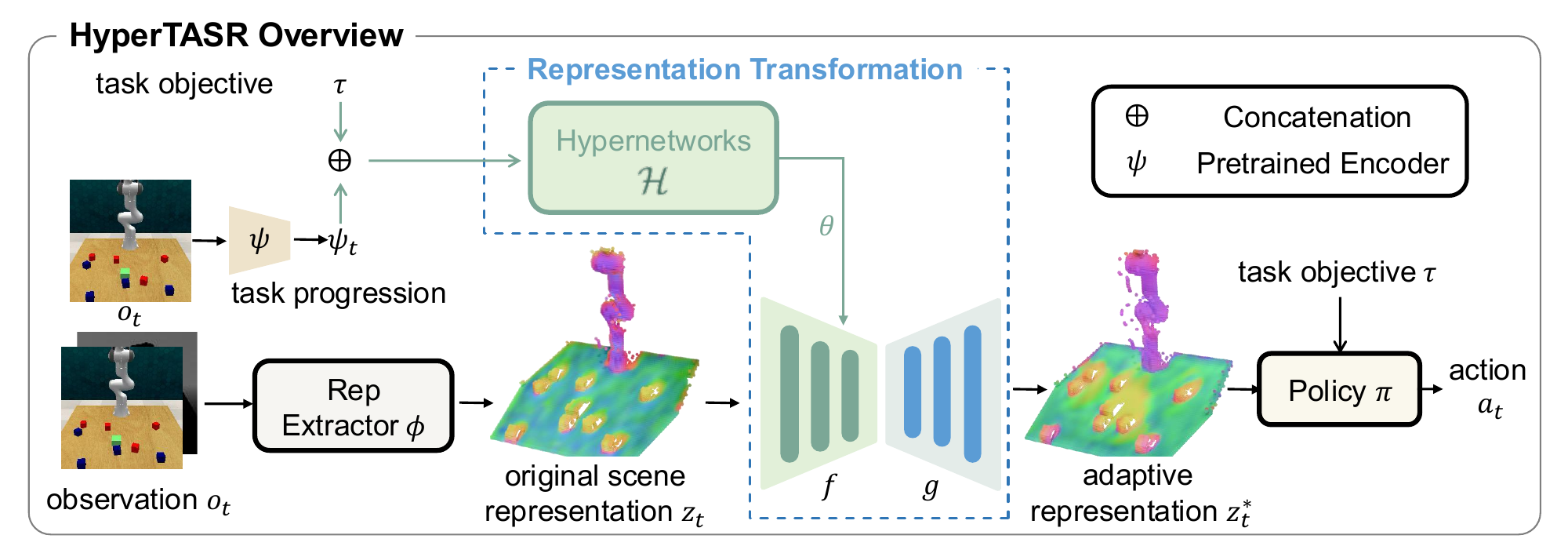}
   \vspace{-5mm}
   \caption{HyperTASR framework overview. 
   Our approach enhances policy learning pipelines by introducing a dynamic scene representation mapping to transform the representation before performing action prediction. 
   The mapping consists of a hypernetwork and task-specific autoencoder (highlighted in blue). 
   The hypernetwork dynamically generates encoder parameters conditioned on both task objectives and progression state, 
   enabling contextual modulation of scene representations throughout task execution.
   }
   \vspace{-3mm}
   \label{fig:overview_method}
\end{figure}

In this section, we present HyperTASR, our novel hypernetwork-based architecture that dynamically modulates scene representation extraction based on task context and execution phase. 
In Sec.~\ref{sec:prelims}, we formalize the problem definition within the context of manipulation policy learning,
followed by a detailed exposition of our task-conditional representation framework (Sec.~\ref{sec:adaptive-rep})
and the associated training methodologies
(Sec.~\ref{sec:hyper-rep} \& Sec.~\ref{sec:integration}).
Please refer to Fig.~\ref{fig:overview_method} 
for an overview of the proposed policy learning pipeline and the core components. 

\subsection{Preliminaries}
\label{sec:prelims}

Multi-task robotic manipulation 
requires agents to act efficiently across a heterogeneous task space, 
denoted by $\mathcal{T} = \{\tau_k\}_{k=1}^K$. 
Each task $\tau$ induces a task-conditioned Markov Decision Process (MDP), 
characterized by states $s_t \in \mathcal{S}$, 
actions $a_t \in \mathcal{A}$, 
and the task specification $\tau \in \mathcal{T}$. 
The primary objective in multi-task policy learning
is to train a policy 
$\pi: \mathcal{S} \times \mathcal{T} \rightarrow \mathcal{A}$ 
that generates optimal action sequences for completing the task.
For embodied agents operating in physical environments, 
the latent state $s_t$ is rarely directly accessible 
due to inherent limitations in environmental perception. 
{\it Consequently,} 
policies usually operate on learned representations 
$z_t = \phi(o_t)$ 
derived from partial observations $o_t$, 
where $\phi$ denotes the representation extraction network ({\bf representation extractor})
that feeds into the subsequent action prediction module ({\bf policy}).
These processes are also known as POMDPs (Partially Observable Markov Decision Processes).
Here, we adopt the term policy to specifically reference the action prediction component of the learning pipeline. 
In the single-view paradigm, 
agents perceive the environment through a single RGB-D image at each timestep $t$, 
where $o_t = (I_t, D_t)$ 
comprises RGB imagery ($I_t$) and corresponding depth data ($D_t$).

\subsection{Task-Aware Scene Representation}
\label{sec:adaptive-rep}

To formalize, 
contemporary frameworks for multi-task robotic learning 
typically implement a two-stage architecture: 
first extracting a scene representation $z_t = \phi(o_t)$ from observation $o_t$, 
then computing actions via a policy module $a_t = \pi(z_t, \tau)$. 
The task information is integrated 
only at the action prediction stage, 
and the representation extraction process $\phi$ remains 
task-agnostic and static 
throughout task execution.

Specifically, 
we can categorize current representation extraction methodologies into two main approaches. 
The {\it first} approach leverages pre-trained foundational models 
to extract semantically rich scene representations~\cite{ke20243d,gervet2023act3d,d3field}. 
While these methods capture general visual semantics effectively, they operate independently of the downstream action prediction objective, potentially extracting features suboptimal for specific manipulation tasks.
The {\it second} approach 
entails training representation extraction architectures from scratch, 
typically co-optimized with the policy network~\cite{Ze2023GNFactor,ma2024contrastive,driess2022reinforcement}, 
yielding representations calibrated for action prediction.
However, these approaches 
typically employ a task-agnostic representation extractor, 
failing to recognize that different manipulation objectives may require selectively emphasizing distinct aspects of the visual scene.


\textbf{Cognitive Inspiration.}
In contrast, 
human visual processing adaptively modulates scene perception 
based on both task objectives and the execution phase. 
Consider the procedural task of preparing tea: 
during the initial object localization phase, coarse spatial awareness suffices; 
however, during the pouring action, visual processing sharpens to capture precise spatial relationships and fine-grained geometric details necessary for successful liquid transfer.
This dynamic perceptual reconfiguration suggests that effective scene representations should evolve contextually throughout task execution.

\textbf{Proposed Approach.}
{\it Consequently,} 
we propose that scene representation extraction 
should be explicitly conditioned on task context, 
modeled as $z_t = \phi(o_t, \tau)$. 
This task-conditional representation framework 
offers several advantages:
(1) enhanced effectiveness across tasks through adaptive environmental encoding,
(2) improved interpretability as representations selectively highlight task-relevant scene elements,
(3) increased computational efficiency by filtering irrelevant environmental information, and
(4) closer alignment with the progressive nature of manipulation tasks.
The following sections detail our hypernetwork-based implementation that dynamically modulates scene representation extraction according to both task identity and execution phase.

\subsection{Hypernetwork-Driven Task-Conditional Scene Representation}
\label{sec:hyper-rep}

Our approach for integrating task and progression awareness into scene representations 
is designed as a versatile framework applicable across diverse policy learning architectures.
Rather than modifying the intrinsic representation extraction mechanisms of each pipeline -- which would introduce 
architectural dependencies and complicate comparative analysis -- we propose a modular transformation layer 
that preserves the original dimensionality of scene representations while enriching them with task-specific context.
{\it Specifically,} 
we implement this transformation as a lightweight autoencoding structure:
\begin{equation}
  z^*_t = g^{\omega} \circ f(z_t; \theta),
\end{equation}
where $f$ and $g$ denote the encoding and decoding functions, respectively, 
parameterized by $\theta$ and $\omega$. 
This formulation ensures that the transformed representation $z^*_t$ maintains the same dimensionality as the original $z_t$, 
enabling seamless integration with any downstream policy network without architectural modifications.

To incorporate contextual information about the task 
into the representation transformation process, 
we dynamically modulate the encoding function parameters $f(\cdot;\theta)$ rather than the features. 
{\it Crucially,} 
we identify two fundamental dimensions that guide adaptive representation:
the {\it task objective} that defines the manipulation goal and the {\it task progression state} that captures temporal execution context. 
We formalize these dimensions with 
task specification $\tau$ and progression encoding 
$\psi_t = \psi(o_t)$ extracted from observation $o_t$. 
The conditional transformation process is expressed as:
\begin{equation}
z^*_t = g^{\omega} \circ f(z_t; \theta(\tau, \psi_t)).
\label{eq:task_encoding}
\end{equation}
While the decoding function 
$g^\omega$ remains task-invariant and is co-optimized during training, 
the scene representation encoder parameters $\theta$ are dynamically generated to adapt to the task context.
{\it For this dynamic parameter generation,} 
we leverage a hypernetwork architecture $\mathcal{H}$ \cite{renhypogen} that synthesizes 
task-specific encoding parameters
conditioned on both task objectives and execution state: 
\begin{equation}
  \theta = \mathcal{H}(\tau, \psi_t).
\end{equation}
This hypernetwork-based parameterization provides three crucial technical advantages:
(1) it establishes a clear computational separation between task-contextual and state-dependent gradient flows during backpropagation~\cite{sarafian2021recomposing}, substantially enhancing learning efficiency;
(2) it enables functional transformation of the representation space rather than mere feature weighting or selection, allowing for more expressive adaptation to task requirements; and
(3) it facilitates rapid adaptation across tasks without catastrophic forgetting, as task-specific parameters are generated on demand.
The resulting architecture dynamically reconfigures its representation extraction strategy for each task and execution phase while maintaining nice compatibility with existing policy learning frameworks.

\subsection{Integration and Training Objectives}
\label{sec:integration}

We integrate our hypernetwork-based task-aware scene representation extraction 
with two representative state-of-the-art architectures that exemplify distinct approaches to representation learning: 
GNFactor~\cite{Ze2023GNFactor}, which trains representations from scratch with the policy, 
and 3D Diffuser Actor~\cite{ke20243d}, which leverages pre-trained visual backbones.

\textbf{GNFactor Integration.}
We insert a 3D autoencoder after GNFactor's volumetric representation extraction, with encoder parameters generated by the proposed HyperTASR. 
{\it It is worth noting that} 
we eliminate the feature distillation component 
consists of neural rendering.
{\it This modification streamlines the framework} to be optimized end-to-end solely through behavior cloning, formulated as:
\begin{equation}
   \mathcal{L} 
   =\lambda_{\text{pos}}\mathcal{L}_{\text{pos}} + \lambda_{\text{rot}}\mathcal{L}_{\text{rot}} +
   \lambda_{\text{open}}\mathcal{L}_{\text{open}} +
   \lambda_{\text{collide}}\mathcal{L}_{\text{collide}},
\end{equation}
where $\mathcal{L}_{\text{pos}}$, $\mathcal{L}_{\text{rot}}$, $\mathcal{L}_{\text{open}}$, and $\mathcal{L}_{\text{collide}}$ represent the position loss, rotation loss, gripper openness loss, and collision avoidance loss, respectively.

\textbf{3D Diffuser Actor Integration.}
For 3D Diffuser Actor, which utilizes point cloud features 
derived from a pre-trained 2D visual backbone, 
we integrate HyperTASR by inserting a 2D autoencoder after 
the pre-trained feature extraction stage. 
This placement allows our hypernetwork to modulate the rich semantic features 
before they are projected into the 3D point cloud representation.
We maintain the original training objectives, 
optimizing the complete architecture through behavior cloning loss.

Our hypernetwork, autoencoder components, and policy networks are jointly optimized through gradient backpropagation. 
This integration approach demonstrates the versatility of our framework, 
as it seamlessly enhances both learned-from-scratch and pre-trained representation architectures 
without requiring a fundamental redesign of their core components.

\vspace{-2mm}
\section{Experiments}
\vspace{-1mm}

We evaluate HyperTASR by integrating it with GNFactor \cite{Ze2023GNFactor} and 3D Diffuser Actor \cite{ke20243d} on multi-task manipulation benchmarks in both simulation and real-world settings.




\definecolor{ImportantColor}{rgb}{0.63, 0.79, 0.95}
\newcommand{\rebut}[1]{{\color{black}#1}}
\newcommand{\tb}[3]{\setlength{\tabcolsep}{#2mm}\begin{tabular}{#1}#3\end{tabular}}

\begin{table*}[t!]
    \centering
    \begin{adjustbox}{width=0.99\textwidth}
    \tb{@{}l|c|c|c|c|c|c|c|c|c|c|c|c@{}}{1.0}{
    \toprule
    & \cellcolor{ImportantColor}Avg. & Avg. & close & open & sweep to & meat off & turn & slide & put in & drag & push & stack \\
    & \cellcolor{ImportantColor}Success $\uparrow$ & Rank $\downarrow$ & jar & drawer & dustpan & grill & tap & block & drawer & stick & buttons & blocks \\
    \midrule
    Peract~\cite{shridhar2023perceiver} &  \cellcolor{ImportantColor}20.4 & 5.4 &  18.7\rebut{$_{\pm 8.2}$} &  54.7\rebut{$_{\pm 18.6}$}  & 0.0\rebut{$_{\pm 0.0}$}  &  40.0\rebut{$_{\pm 17.0}$} &  38.7\rebut{$_{\pm 6.8}$} & 18.7\rebut{$_{\pm 13.6}$}  & 2.7\rebut{$_{\pm 3.3}$}  & 5.3\rebut{$_{\pm 5.0}$}  & 18.7\rebut{$_{\pm 12.4}$}  &  6.7\rebut{$_{\pm 1.9}$} \\
    
    GNFactor~\cite{Ze2023GNFactor} &  \cellcolor{ImportantColor}33.3 & 4.8 &  32.8\rebut{$_{\pm 0.6}$} &  36.0\rebut{$_{\pm 0.2}$}  & 48.0\rebut{$_{\pm 0.3}$}  &  51.2\rebut{$_{\pm 0.1}$} &  56.8\rebut{$_{\pm 0.3}$} & 20.0\rebut{$_{\pm 1.2}$}  & 8.8\rebut{$_{\pm 1.3}$}  & 69.6\rebut{$_{\pm 0.4}$}  & 5.6\rebut{$_{\pm 3.7}$}  &  4.0\rebut{$_{\pm 4.0}$} \\

    Act3D~\cite{gervet2023act3d} & \cellcolor{ImportantColor}65.3 & 3.1 & 52.0\rebut{$_{\pm 5.7}$} &  84.0\rebut{$_{\pm 8.6}$}  & 80.0\rebut{$_{\pm 9.8}$} & 66.7\rebut{$_{\pm 1.9}$} & 64.0\rebut{$_{\pm 5.7}$} & \textbf{100.0}\rebut{$_{\pm 0.0}$} & 54.7\rebut{$_{\pm 3.8}$} & 86.7\rebut{$_{\pm 1.9}$} &  64.0\rebut{$_{\pm 1.9}$} &  0.0\rebut{$_{\pm 0.0}$} \\

    3D Diffuser Actor~\cite{ke20243d} & \cellcolor{ImportantColor}79.0 & 1.8 &  63.2\rebut{$_{\pm 1.6}$} &  \textbf{88.8}\rebut{$_{\pm 7.8}$}  & 94.4\rebut{$_{\pm 4.1}$}  &  \textbf{84.8}\rebut{$_{\pm 4.7}$} &  72.8\rebut{$_{\pm 4.7}$} & 94.4\rebut{$_{\pm 2.0}$}  & 88.8\rebut{$_{\pm 4.7}$}  & 98.4\rebut{$_{\pm 2.0}$}  & 87.2\rebut{$_{\pm 1.6}$}  &  \textbf{17.4}\rebut{$_{\pm 5.1}$} \\
    
    \midrule
    GNFactor w/ HyperTASR &  \cellcolor{ImportantColor}42.6 & 4.5 & 32.0\rebut{$_{\pm 0}$} &  
    75.2\rebut{$_{\pm 0.5}$}  & 66.4\rebut{$_{\pm 0.4}$}  &  48.8\rebut{$_{\pm 0.2}$} &  54.4\rebut{$_{\pm 2.0}$} & 23.2\rebut{$_{\pm 4.6}$}  & 22.4\rebut{$_{\pm 0.9}$}  & 83.2\rebut{$_{\pm 0.3}$}  & 17.6\rebut{$_{\pm 1.2}$}  &  3.2\rebut{$_{\pm 3.5}$} \\
    
    3D DA w/ HyperTASR &  \cellcolor{ImportantColor}\textbf{81.3} & \textbf{1.4} & \textbf{68.0}\rebut{$_{\pm 2.5}$} &  87.2\rebut{$_{\pm 1.6}$}  & \textbf{98.4}\rebut{$_{\pm 2.0}$} & 82.4\rebut{$_{\pm 3.2}$}  &  \textbf{85.6}\rebut{$_{\pm 3.2}$} &  98.4\rebut{$_{\pm 2.0}$}  & \textbf{89.6}\rebut{$_{\pm 6.5}$}  & \textbf{100.0}\rebut{$_{\pm 0.0}$} & \textbf{92.0}\rebut{$_{\pm 0.0}$}  &  11.2\rebut{$_{\pm 1.6}$} \\
    \bottomrule
    }
    \end{adjustbox}
    \caption{\textbf{Evaluation on RLBench in the single-view setting.} Success rates on 10 RLBench tasks using only the {\it front} camera view. All models are trained with 20 demonstrations per task and evaluated across 5 seeds with 25 episodes per task. HyperTASR significantly improves performance when integrated with both GNFactor~\cite{Ze2023GNFactor} and 3D Diffuser Actor (3D DA)~\cite{ke20243d}, demonstrating the effectiveness of task-aware scene representations.}
    \vspace{-3mm}
    \label{tab:main_res}
\end{table*}

\subsection{Experiment Setting}
Our experiments use RLBench \cite{james2020rlbench}, a large-scale benchmark with over 100 manipulation tasks in realistic simulated environments using a Franka Panda robot. Following \cite{Ze2023GNFactor}, we evaluate on 10 language-conditioned tasks comprising 166 variations. All methods predict the next keypose for the end-effector and use BiRRT \cite{kuffner2000rrt} for motion planning. To emphasize the impact of scene representations under practical deployment constraints, we also conduct all experiments in the challenging single-view setting, using only the front camera view RGB-D sensory data.

\textbf{Baselines.} We compare against state-of-the-art policy learning frameworks: PerAct \cite{shridhar2023perceiver}, which voxelizes the 3D workspace; GNFactor \cite{Ze2023GNFactor}, which constructs a 3D feature volume from a single RGB-D view; and Act3D \cite{gervet2023act3d} and 3D Diffuser Actor \cite{ke20243d}, which represent the state of the art on RLBench. Results for PerAct and Act3D are adopted from published work \cite{Ze2023GNFactor,ke20243d}, while we retrained GNFactor and 3D Diffuser Actor under identical single-view conditions to ensure fair comparison. All models are trained on the same keypose demonstrations, and we report results across five random seeds to ensure statistical reliability.

\subsection{Implementation Details}

Our HyperTASR implementation uses a UNet-based~\cite{ronneberger2015u} autoencoder with skip connections. 
Following~\cite{renhypogen}, we employ an optimization-biased hypernetwork that predicts parameter updates iteratively rather than 
directly generating encoder weights 
via fully connected layers. 
For task objective conditioning ($\tau$), we utilize the language features already present in the original policy pipelines. Task progression information ($\psi_t$) is extracted using a frozen pretrained VAE Encoder from Stable Diffusion~\cite{rombach2022high}.
For GNFactor integration, we directly apply our HyperTASR to predict parameters of the original lightweight 3D UNet voxel encoder. 
The model is trained for 200k iterations on a single NVIDIA H800 GPU. For 3D Diffuser Actor, we maintain the fixed backbone and add a 2D UNet with nine convolutional layers, training for 600k iterations on four H800 GPUs.
\begin{figure*}[!t]
    \centering
    \vspace{-3mm}
    \includegraphics[width=0.93\linewidth]{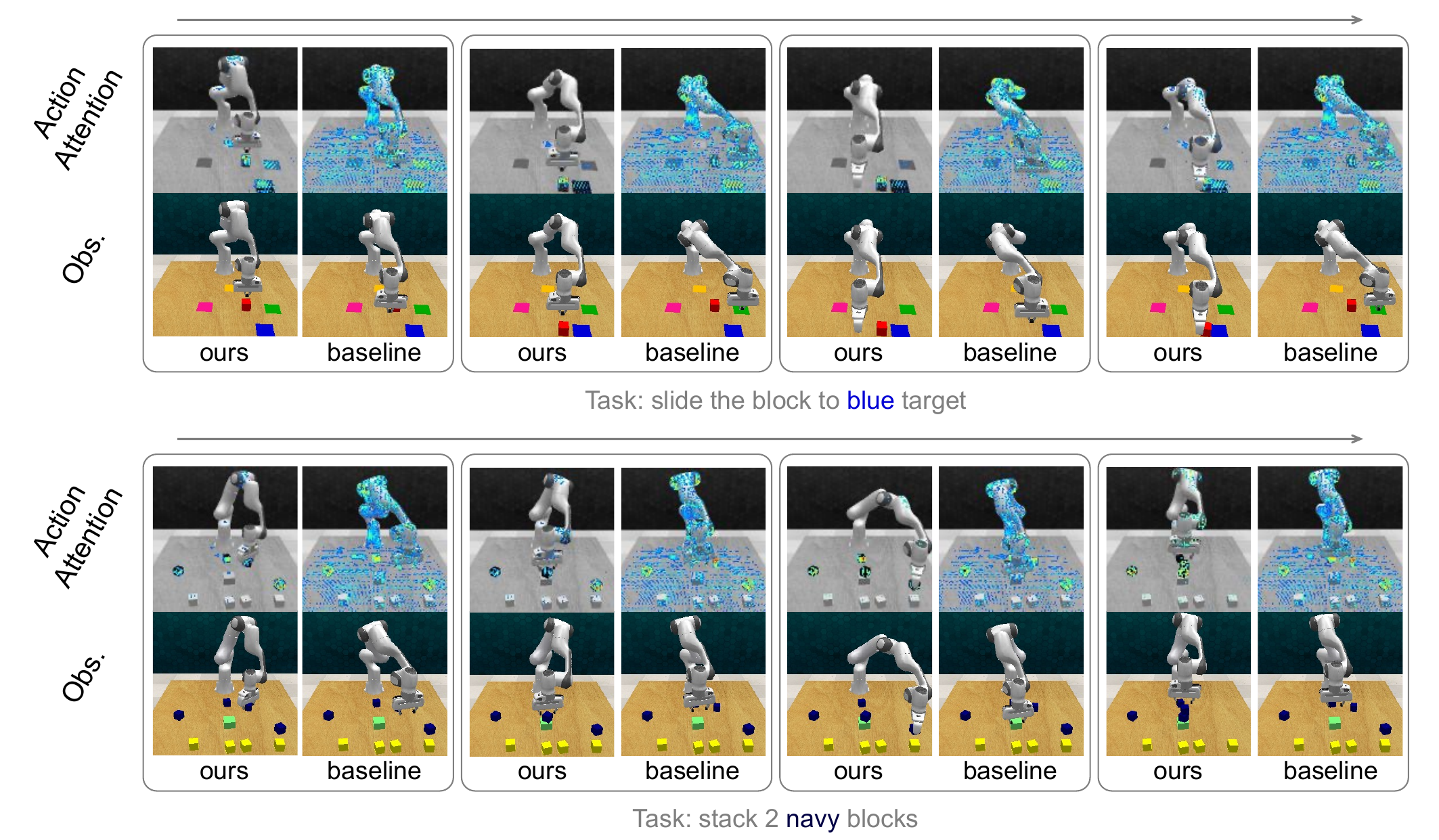}
    \vspace{-3mm}
    \caption{Attention visualization comparing policies with and without task-aware scene representations on `slide block' (top) and `stack blocks' (bottom) tasks. HyperTASR consistently focuses on task-relevant objects, while the baseline attention disperses across irrelevant scene elements, explaining the performance gain of HyperTASR across approaches.
    }
    \vspace{-3mm}
   \label{fig:evidence-of-effectiveness}
\end{figure*}
\subsection{Results in Simulation}
We present quantitative results in Table~\ref{tab:main_res}, evaluating performance through both average success rate and average rank across tasks. The experimental findings demonstrate HyperTASR's substantial impact on both integration frameworks.
When integrated with GNFactor, HyperTASR achieves a remarkable 27.9\% relative improvement (9.3\% absolute increase) over the baseline across the 10 evaluation tasks. Notably, this performance gain comes with reduced computational requirements -- our approach eliminates the need for feature distillation, removes multi-view supervision dependencies, and results in a smaller network with faster training convergence. This combination of enhanced performance with reduced computational overhead highlights the efficiency of task-aware scene representations.
For the 3DDA integration, HyperTASR surpasses not only the baseline but all current state-of-the-art methods, establishing new benchmark performance for single-view manipulation. The consistent improvement across both frameworks demonstrates HyperTASR's versatility and effectiveness with different scene representation paradigms and policy network architectures.
The average rank metric further confirms that models enhanced with HyperTASR perform better across a wider range of tasks, indicating superior cross-task generalization capabilities -- a critical attribute for real-world robotic applications in diverse manipulation scenarios.

\paragraph{Visualization of HyperTASR.} 
To provide qualitative insights into how task-aware scene representations transform attentional dynamics, we visualize action attention patterns in Fig.~\ref{fig:evidence-of-effectiveness}. Compared to baseline models, HyperTASR produces significantly more focused attention maps that precisely target task-relevant objects. 
Furthermore, these attention patterns evolve dynamically throughout task execution. In the sliding blocks task, for example, attention initially concentrates on the block while the gripper approaches. 
Once contact is established, attention shifts to the target area where the block should be placed. 
This progressive adaptation of perceptual focus closely mirrors human visual processing during manipulation tasks, providing a clear mechanism for the performance improvements observed in our quantitative results.
\vspace{-1mm}

\subsection{Real-World Evaluation}

To validate HyperTASR's effectiveness in physical environments, we conduct experiments using a Piper robotic arm equipped with a parallel gripper. We design six diverse manipulation tasks with variations in object colors, counts, placements, and categories to assess generalization capabilities. For each task, we collect 15 expert demonstrations using a master-puppet teleoperation system identical to ALOHA~\cite{fu2024mobile}.
RGB-D observations are captured via an Intel RealSense camera at 640×480 resolution and subsequently downsampled to 256×256 for processing. During inference, target gripper poses are executed using the MoveIt package in ROS. We integrate HyperTASR with the 3D Diffuser Actor framework and evaluate performance across 15 episodes per task.

\definecolor{ImportantColor}{rgb}{0.63, 0.79, 0.95}

\begin{wraptable}{r}{0.70\textwidth}
    \centering
    \vspace{-4mm}
    \begin{adjustbox}{width=0.7\textwidth}
    \begin{tabular}{c|c|c|c|c|c|c|c}
        \toprule
        & \cellcolor{ImportantColor}Avg.   & place  & clean  & stack  & stack  & put cups  & place \\
        & \cellcolor{ImportantColor}Succ  & dish  & cups  & cups  & blocks  & on shelf  & blocks  \\
        \midrule
        3D Diffuser Actor & \cellcolor{ImportantColor}42.2 & 40.0 & 53.3 & 13.3 & 20.0 & 46.6 & 80.0 \\
        \textbf{3D-DA w/ HyperTASR} & \cellcolor{ImportantColor}\textbf{51.1} & \textbf{53.3} & \textbf{66.6} & \textbf{20.0} & \textbf{26.6} &  \textbf{53.3} & \textbf{86.6} \\
        \bottomrule
    \end{tabular}
    \end{adjustbox}
    \vspace{-3mm}
    \caption{\textbf{Real-World Experiment Results.} Success rates across six manipulation tasks with 15 episodes per task. HyperTASR consistently outperforms the baseline 3D Diffuser Actor, demonstrating that task-aware scene representations transfer effectively from simulation to physical environments with limited demonstration data.}
    \vspace{-4mm}
    \label{tab:real}
\end{wraptable}

As shown in Table~\ref{tab:real}, HyperTASR consistently outperforms the baseline 3D Diffuser Actor across all real-world tasks, demonstrating that the benefits of task-aware scene representations transfer effectively from simulation to physical environments. This performance gain is particularly noteworthy given the limited demonstration data (15 per task) and the inherent challenges of real-world sensing and actuation. Additional visualizations and experimental details are provided in the Appendix.

\begin{wraptable}{r}{0.50\textwidth}
    \centering
    \vspace{-4mm}
    \begin{adjustbox}{width=0.5\textwidth}
    \begin{tabular}{cc}
        \toprule
        Ablation & Success Rate (\%) \\
        \midrule
        3D Diffuser Actor & 79.02 $\pm$ 1.65\\
        Task-Awareness by Transformer & 79.23 $\pm$ 1.10\\
        \textbf{Task-awareness by HyperTASR (ours)} & \textbf{81.28 $\pm$ 0.82} \\
        \midrule
        GNFactor & 33.20 $\pm$ 1.22 \\
        HyperTASR w/ Feature Distillation & 34.00 $\pm$ 2.12 \\
        HyperTASR conditioned on $\tau$ & 32.24 $\pm$ 0.60\\
        HyperTASR predicting $\theta$ and $\omega$ & 36.32 $\pm$ 1.32\\
        \textbf{HyperTASR (ours)} & \textbf{42.60 $\pm$ 1.35} \\
        \bottomrule
    \end{tabular}
    \end{adjustbox}
    \vspace{-2mm}
    \caption{\textbf{Ablation Study Results.} {\it Upper:} Comparison of hypernetwork vs. attention-based approaches for implementing task awareness in the 3D Diffuser Actor framework. 
    {\it Lower:} Analysis of feature distillation, task progression conditioning, and hypernetwork target selection within the GNFactor framework.}
    \vspace{-4mm}
    \label{tab:ablation_table}
\end{wraptable}
\vspace{-2mm}
\subsection{Ablation Study}

We conduct comprehensive ablation studies to evaluate key design choices in HyperTASR.
{\it First,} we examine whether simpler conditioning mechanisms could achieve similar benefits. Within the 3D Diffuser Actor framework, we replace our hypernetwork with a cross-attention module that fuses task objectives and progression information with the original scene representation. As shown in Table~\ref{tab:ablation_table} (upper), this alternative yields only marginal improvements, confirming that effective task-aware representations require sophisticated functional transformation rather than simple feature fusion.
We {\it further} investigate three critical aspects using GNFactor, with results in Table~\ref{tab:ablation_table} (lower):
(i) \textbf{Feature Distillation:} Adding explicit distillation supervision constrains representational flexibility, reducing performance by limiting adaptation capabilities -- supporting our design choice to eliminate this component.
(ii) \textbf{Task Progression:} Conditioning only on task objectives ($\tau$) without progression information significantly degrades performance, confirming that effective representations must evolve throughout task execution.
(iii) \textbf{Hypernetwork Target:} Having the hypernetwork predict only encoder parameters proves more efficient than generating the entire autoencoder, validating our architectural focus on encoder transformation with a fixed decoder.

\vspace{-1mm}
\section{Conclusion}
\label{sec:conclusion}

We present HyperTASR, 
a novel framework for task-aware scene representations in robotic manipulation that dynamically adapts perceptual processing 
based on both task objectives and execution progression. 
Our hypernetwork-driven approach 
enables representations to evolve contextually throughout task execution, 
focusing on task-relevant environmental features.
Evaluations in both simulation and real-world settings demonstrate significant performance enhancements across different representation paradigms.
Ablation studies confirm the effectiveness of our design components for task-aware representations extraction.
HyperTASR bridges the gap between human-inspired adaptive perception and computational approaches to robotic manipulation, establishing a foundation for more efficient multi-task policy learning.

\section{Limitations}

While HyperTASR demonstrates substantial improvements in manipulation performance, several opportunities for future enhancement remain. Our experiments primarily focus on behavior cloning, while HyperTASR has the capability of extending to reinforcement learning field.

Our current evaluation uses single-arm grippers as the robotic platform. The principles of task-aware scene representation could potentially extend to more advanced manipulation systems such as bimanual setups and dexterous hands, which would broaden the applicability of our approach to more sophisticated manipulation tasks.

These limitations highlight promising research directions that could build upon the foundation established by HyperTASR. The consistent performance improvements observed across different representation paradigms suggest that task-aware adaptation principles could generalize effectively to these extended capabilities.

\acknowledgments{
This work is supported by 
the Early Career Scheme of the Research Grants Council (RGC) grant \# 27207224,
the HKU-100 Award, 
a donation from the Musketeers Foundation, 
and in part by the JC STEM Lab of Autonomous Intelligent Systems funded by The Hong Kong Jockey Club Charities Trust.
}


\bibliography{main}  

\clearpage
\appendix

\section{Additional Implementation Details}

\subsection{Model Structure}


\begin{figure*}[!h]
  \centering
  \vspace{-1em}
   \includegraphics[width=\linewidth]{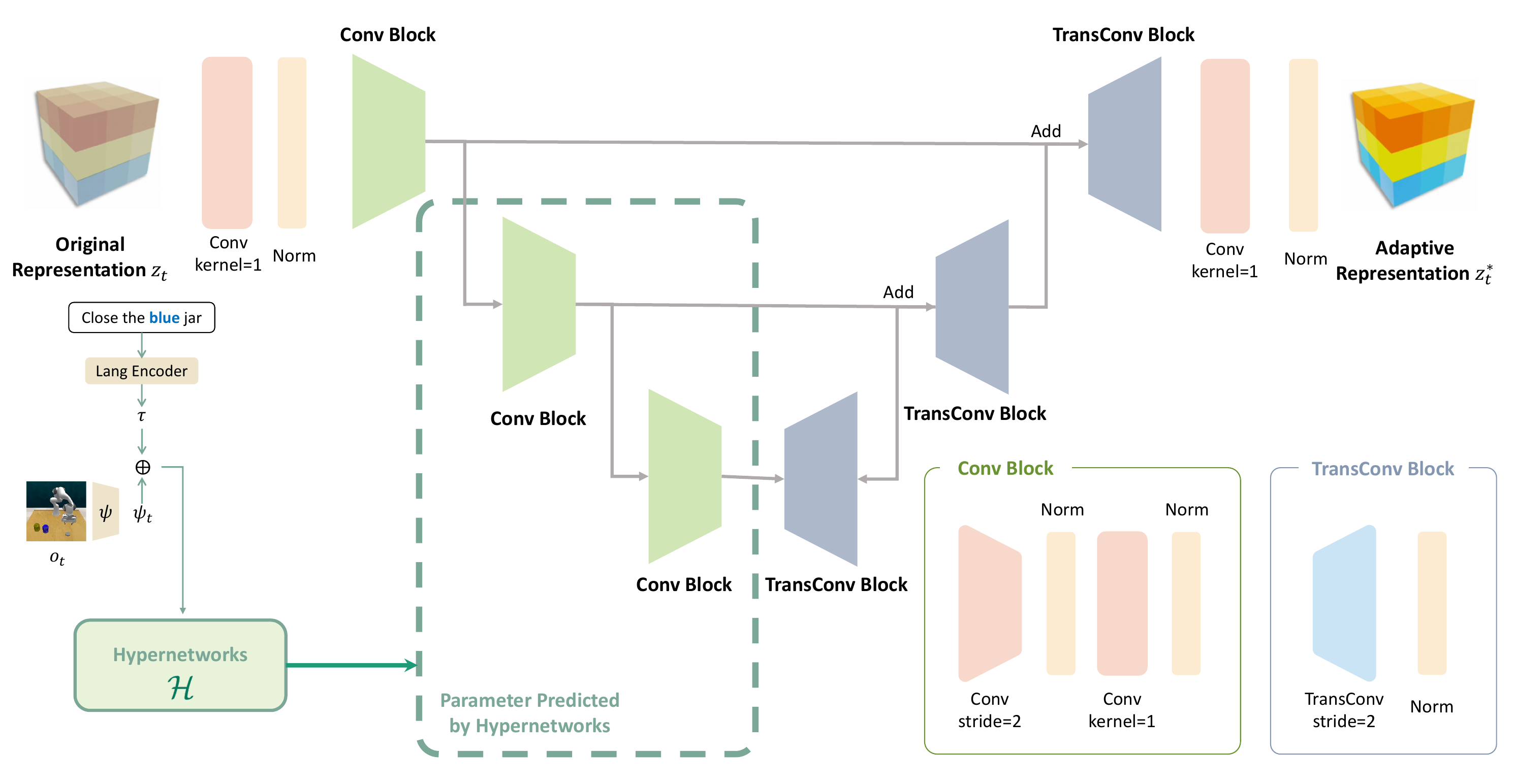}
   \caption{\textbf{The detailed model structure of our HyperTASR.} The main diagram shows that the pipeline consists of 2 Convolutional layer and a UNet with skip-connection. {\it Bottom-right:} The detailed structure of \textbf{Conv Block} and \textbf{TransConv Block} in the pipeline. These blocks serve as the key components for the encoding and decoding process.}
   \label{fig:adpt_struct}
\end{figure*}

HyperTASR we design mainly consists of three convolutional blocks and three transposed convolutional blocks, as detailed in Fig.~\ref{fig:adpt_struct}. Each convolutional block contains two convolutional layers, followed by an InstanceNorm layer and a Leaky ReLU activation function. The first convolutional layer in each block has a kernel size of 3 and a stride of 2, which reduces the resolution of the feature map while increasing the feature channel dimension, effectively encoding the features. The second convolutional layer has a kernel size of 1 and does not change the resolution or channel dimension of the feature map, serving to refine the encoded features.

The transposed convolutional blocks are relatively simpler, consisting of a single transposed convolutional layer followed by InstanceNorm and Leaky ReLU activation. The transposed convolutional layer increases the resolution of the feature map while reducing the channel dimension, effectively decoding the features. This layer has a kernel size of 3 and a stride of 2, ensuring that the spatial dimensions of the feature map are expanded appropriately.

HyperTASR used in GNFactor~\cite{Ze2023GNFactor} and 3D Diffuser Actor~\cite{ke20243d} follow the aforementioned structure. GNFactor directly utilizes a 3D deep volume as its representation, requiring 3D convolutions and 3D transposed convolutions. For the 3D Diffuser Actor, the representation is a point cloud feature, which in a single-view setup combines a 2D feature map and a depth map, leading us to employ 2D convolutions and 2D transposed convolutions as the core elements of the UNet architecture.

\begin{figure*}[!h]
  \centering
  \vspace{-1em}
   \includegraphics[width=\linewidth]{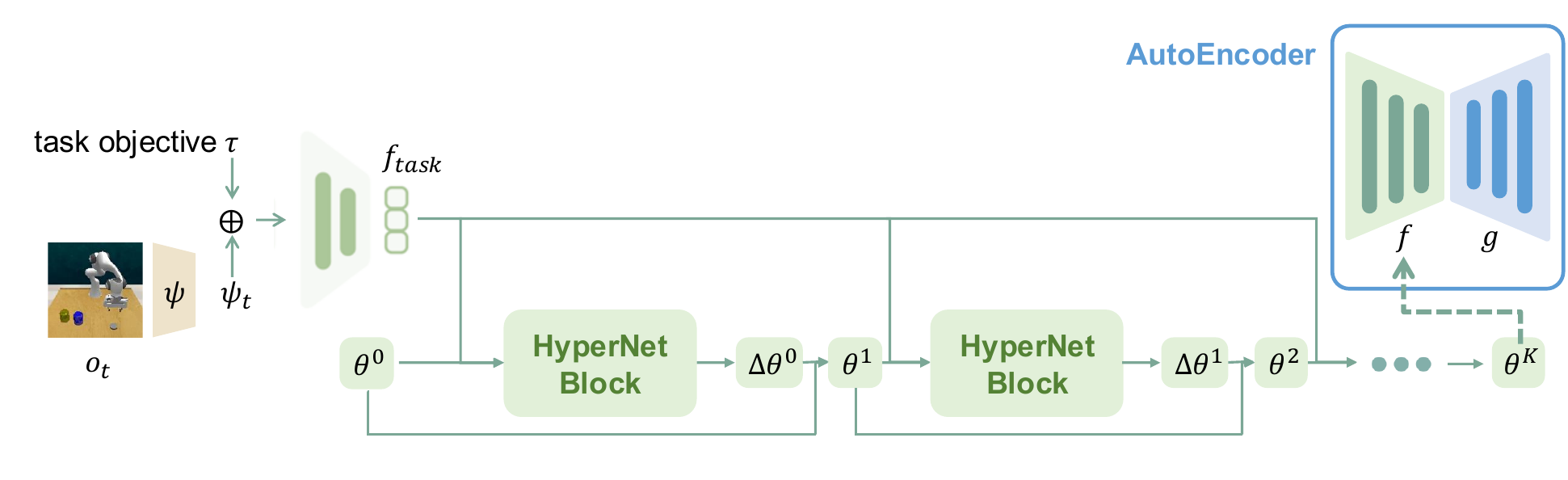}
   \caption{\textbf{The detailed structure of hypernetworks we used in HyperTASR.} We employ an optimization-biased hypernetwork that predicts parameter updates iteratively rather than 
directly generating encoder weights 
via fully connected layers. }
   \label{fig:hyper_structure}
\end{figure*}

In Fig.~\ref{fig:hyper_structure}, we detail the implementation of our hypernetworks. Following~\cite{renhypogen}, we adopt an optimization-based hypernetwork, which iteratively predicts the parameter updates rather than directly predicting the final parameter.
In our implementation, $K=8$ represents the parameter update iteration.
To control the parameter size of the UNet and effectively manage the parameter size of the Hypernetworks, we introduce optional encoders and decoders. The encoder reduces the dimensionality of the input features before they are fed into the UNet, while the decoder restores the dimensionality after processing. This mechanism is particularly useful for maintaining a balance between model complexity and performance. Specifically, for the 3D Diffuser Actor, we incorporate these optional encoders and decoders to better adapt to the varying input feature requirements.

\subsection{Dataset Composition}
\begin{table*}[htbp]
\centering
\begin{adjustbox}{width=0.99\textwidth}
\begin{tabular}{lcccl}
\toprule
Task & Variation Type & \# of Variations & Avg. Keyframes & Language Description Example \\ 
\midrule
\texttt{close jar} & color & 20 & 6.0 & ``close the --- jar" \\

\texttt{meat off grill} & category & 2 & 5.0 & ``take the --- off the grill" \\

\texttt{open drawer} & placement & 3 & 3.0 & ``open the --- drawer\\

\texttt{sweep to dustpan} & size & 2 & 4.6 & ``sweep dirt to the --- dustpan" \\

\texttt{turn tap} & placement & 2 & 2.0 & ``turn --- tap"\\

\texttt{slide block} & color &  4 & 4.7 & ``slide the block to --- target” \\

\texttt{put in drawer} & placement & 3 & 12.0 & ``put the item in the --- drawer” \\

\texttt{drag stick} & color & 
20 & 6.0 &  ``use the stick to drag the cube onto the --- --- target”\\

\texttt{push buttons} & color & 50 &3.8 & ``push the --- button, [then the --- button]”
\\

\texttt{stack blocks} & color, count & 60 & 14.6 & ``stack --- --- blocks” \\

\bottomrule
\end{tabular}
\end{adjustbox}
\caption{\textbf{Dataset composition of 10 manipulation tasks in RLBench~\cite{james2020rlbench}.}}
\label{tab:task_desc}
\vspace{-1em}
\end{table*}

\begin{table*}[htbp]
\centering
\begin{tabular}{lcccl}
\toprule
Task & Variation Type & Language Description Example \\ 
\midrule
\texttt{place dish} & color & ``place the --- dish on the --- tablecloth" \\

\texttt{clean cups} & color, placement & ``Put the --- cup into the --- basket" \\

\texttt{stack cups} & color, placement & ``stack the --- cup on the --- cup\\

\texttt{stack blocks} & color, count & ``stack the --- cubes" \\

\texttt{put cups on shelf} & placement & ``put the --- cup on the shelf next to --- cup"\\

\texttt{place blocks} & color & ``Place the --- block on the --- plate” \\

\bottomrule
\end{tabular}
\caption{\textbf{Dataset composition of 6 manipulation tasks in real robot experiments.}}
\label{tab:real_desc}
\vspace{-1em}
\end{table*}
We conduct experiments on 10 language-conditioned manipulation tasks from RLBench~\cite{james2020rlbench}, which align with the experimental setup of GNFactor~\cite{Ze2023GNFactor}. The task variations include randomly sampled attributes such as colors, sizes, counts, placements, and object categories. Detailed descriptions of the variation types, variation numbers, average keyframes, and sample language descriptions for these tasks are provided in Tab.~\ref{tab:task_desc}.

For real robot experiments, we design 6 tasks that cover diverse tasks for "pick and place". We give our sample task description in Tab.~\ref{tab:real_desc}.  
\subsection{Hyperparamters}

\begin{table}[htbp]

\centering
\begin{tabular}{cccccc}
\toprule
Variable Name & Value  \\ 
\midrule
training iteration & $200$k \\
image size & $128\times 128 \times 3$\\
batch size & $1$ \\
optimizer & LAMB\\
learning rate & $0.0005$ \\
input voxel size & $100\times 100 \times 100$\\
number of transformer blocks & $6$ \\
number of latents in PerceiverI/O & $2048$\\
dimension of CLIP language features & $512$\\
\bottomrule
\end{tabular}
\caption{\textbf{Hyperparameters} in GNFactor~\cite{Ze2023GNFactor} Framework.}
\label{tab:hyperparam1}
\end{table}

\begin{table}[htbp]
    \centering
    \tb{@{}cc@{}}{2.0}{
    \toprule
    Variable Name & Value \\
    \midrule
    training iteration & $800k$ \\
    image size & $256\times256\times3$ \\
    batch size & $240$ \\
    optimizer & Adam \\
    learning rate & 0.0001 \\
    embedding dim & $120$ \\
    diffusion timestep & $100$ \\
    loss weight of position and rotaion & $30:20$ \\
    maximal \# of keyposes & $25$ \\
    \bottomrule
    }
    \caption{\textbf{Hyperparameters} in 3D Diffuser Actor~\cite{ke20243d} Framework.}
    \label{tab:hyperparm2}
\end{table}

We provide detailed hyperparameters for our experiments in Tab.~\ref{tab:hyperparam1} and Tab.~\ref{tab:hyperparm2}, with some parallel settings and input data differences compared to GNFactor~\cite{Ze2023GNFactor} and 3D Diffuser Actor~\cite{ke20243d}. To ensure a fair comparison, we reproduce the experiments using the same hyperparameters as the original codebase and report the corresponding results in Tab.~\ref{tab:main_res}. These results serve as a benchmark for understanding the impact of our modifications.

\paragraph{Impact of Hyperparameter Changes on Experimental Results.} For the GNFactor framework, we opt not to use distributed data parallel (DDP) training. Instead, we utilize a single GPU, halve the batch size, and double the number of training steps. Despite this adjustment, the final reproduced results fall within the range of multiple experimental outcomes reported in~\cite{Ze2023GNFactor}. For the 3D Diffuser Actor, we train both our modified pipeline and the original codebase with the training data provided in the author's released repository, using an RGB image resolution of $256\times256$. Due to the lower resolution compared to the original paper ($256\times256$)~\cite{ke20243d}, our reproduced results (77.0\%) are slightly below the original results (78.4\%). Additionally, slight adjustments to the loss weights are made to account for the resolution difference, and the best configuration is chosen as the unified hyperparameter setting for all our experiments.


\subsection{Computation Cost}

\begin{wraptable}{r}{0.60\textwidth}
    \centering
    \begin{adjustbox}{width=0.99\linewidth}
    \begin{tabular}{ccc}
        \toprule
         & Model Params & Training Time for 1k steps (s) \\
        \midrule
        GNFactor & 64.66M & 976.5 \\
        Adapter (ours) & 99.12M & 844.8 \\
        \bottomrule
    \end{tabular}
    \end{adjustbox}
    \caption{\textbf{Computation Cost.} }
    \label{tab:time}
\end{wraptable}


We calculate the computation cost of our experiments on GNFactor~\cite{Ze2023GNFactor} by measuring both the total number of parameters in the network and the training time, as shown in Tab.~\ref{tab:time}. For training time, we use an unloaded GPU to train for 1k steps and record the time taken. From the results, we observe that while our network has more parameters, it achieves higher training efficiency. The increased parameter count results from the inclusion of Hypernetworks, but this does not negatively impact training efficiency. On the contrary, by removing the neural renderer used for feature distillation, the overall training time is reduced. This demonstrates that HyperTASR does not impose a significant computational burden on the network and, in some cases, even improves efficiency by eliminating certain supervisory components.

\section{Additional Results and Analysis}

\subsection{Additional Ablations Results}

In Tab.~\ref{tab:ablation_table}, we only present the success rate data for the ablation studies on the GNFactor~\cite{Ze2023GNFactor} framework. Here, we further provide detailed results of these ablation studies across all 10 tasks in Tab.~\ref{tab:gnf_ablation}. Additionally, we conduct ablation experiments on the 3D Diffuser Actor, and the corresponding results are shown in Tab.~\ref{tab:3dda_ablation}. These results can validate the effectiveness of our design of HyperTASR.


\begin{table*}[htbp]
    \centering
    \begin{adjustbox}{width=0.99\textwidth}
    \begin{tabular}{cccccccccccc}
        \toprule
        & Avg.   & close  & open  & sweep to  & meat off  & turn  & slide  & put in\  & drag  & push  & stack \\
        & Success  & jar  & drawer  & dustpan  & grill  & tap  & block  & drawer  & stick  & buttons  & blocks \\
        \midrule
         
        GNFactor & 33.3 & \textbf{32.8} & 36.0 & 48.0 & 51.2 & \textbf{56.8} & 20.0 & 8.8 & 69.6 & 5.6 & 4.0\\

        HyperTASR w/ Feature Distillation & 34.0 & 29.6 & 69.6 & 35.2 & 50.4 & 44.0 & 8.0 & 1.6 & 48.6 & \textbf{43.2} & 8.8\\

        HyperTASR conditioned on $\tau$ & 32.2 & 10.4 & 47.2 & 29.6 & 34.4 & 51.2 & 16.8 & 10.4 & \textbf{92.0} & 22.4 & 8.0\\

        HyperTASR predicting $\theta$ and $\omega$ & 36.3 & 28.0 & 67.2 & 20.0 & \textbf{60.8} & 49.6 & 20.0 & 18.4 & 78.4 & 7.2 & \textbf{13.6}\\
        
        \textbf{GNFactor w/ HyperTASR} & \textbf{42.6} & 32.0 & \textbf{75.2} & \textbf{66.4} & 48.8 & 54.4 & \textbf{23.2} & \textbf{22.4} & 83.2 & 17.6 & 3.2 \\
        \bottomrule
    \end{tabular}
    \end{adjustbox}
    \caption{\textbf{Detailed Ablation Study Results in GNFactor framework}. 
    We report the average success rate 
    across 5 evaluation seeds. }
    \label{tab:gnf_ablation}
\end{table*}

\begin{table*}[htbp]
    \centering
    \begin{adjustbox}{width=0.99\textwidth}
    \begin{tabular}{cccccccccccc}
        \toprule
        & Avg.   & close  & open  & sweep to  & meat off  & turn  & slide  & put in\  & drag  & push  & stack \\
        & Success  & jar  & drawer  & dustpan  & grill  & tap  & block  & drawer  & stick  & buttons  & blocks \\
        \midrule
         
        3DDA & 79.0 & 63.2 & \textbf{88.8} & 94.4 & \textbf{84.8} & 72.8 & 94.4 & 88.8 & 98.4 & 87.2 & \textbf{17.4}\\

        HyperTASR conditioned on $\tau$ & 75.4 & 60.8 & 75.4 & 88.8 & 82.4 & 59.2 & 83.2 & 80.8 & 89.6 & 83.2 & 4.6\\

        HyperTASR predicting $\theta$ and $\omega$ & 79.2 & 66.4 & 86.4 & 96.8 & 81.4 & 79.8 & 84.0 & 87.2 & 99.2 & 89.6 & 10.4\\
        
        \textbf{3DDA w/ HyperTASR} & \textbf{81.3} & \textbf{68.0} & 87.2 & \textbf{98.4} & 82.4 & \textbf{85.6} & \textbf{98.4} & \textbf{89.6} &\textbf{100.0} & \textbf{92.0} & 11.2 \\
        \bottomrule
    \end{tabular}
    \end{adjustbox}
    \caption{\textbf{Detailed Ablation Study Results in 3D Diffuser Actor (3DDA) framework}. 
    We report the average success rate 
    across 5 evaluation seeds. }
    \label{tab:3dda_ablation}
\end{table*}

\begin{figure*}[htbp]
  \centering
   \includegraphics[width=\linewidth]{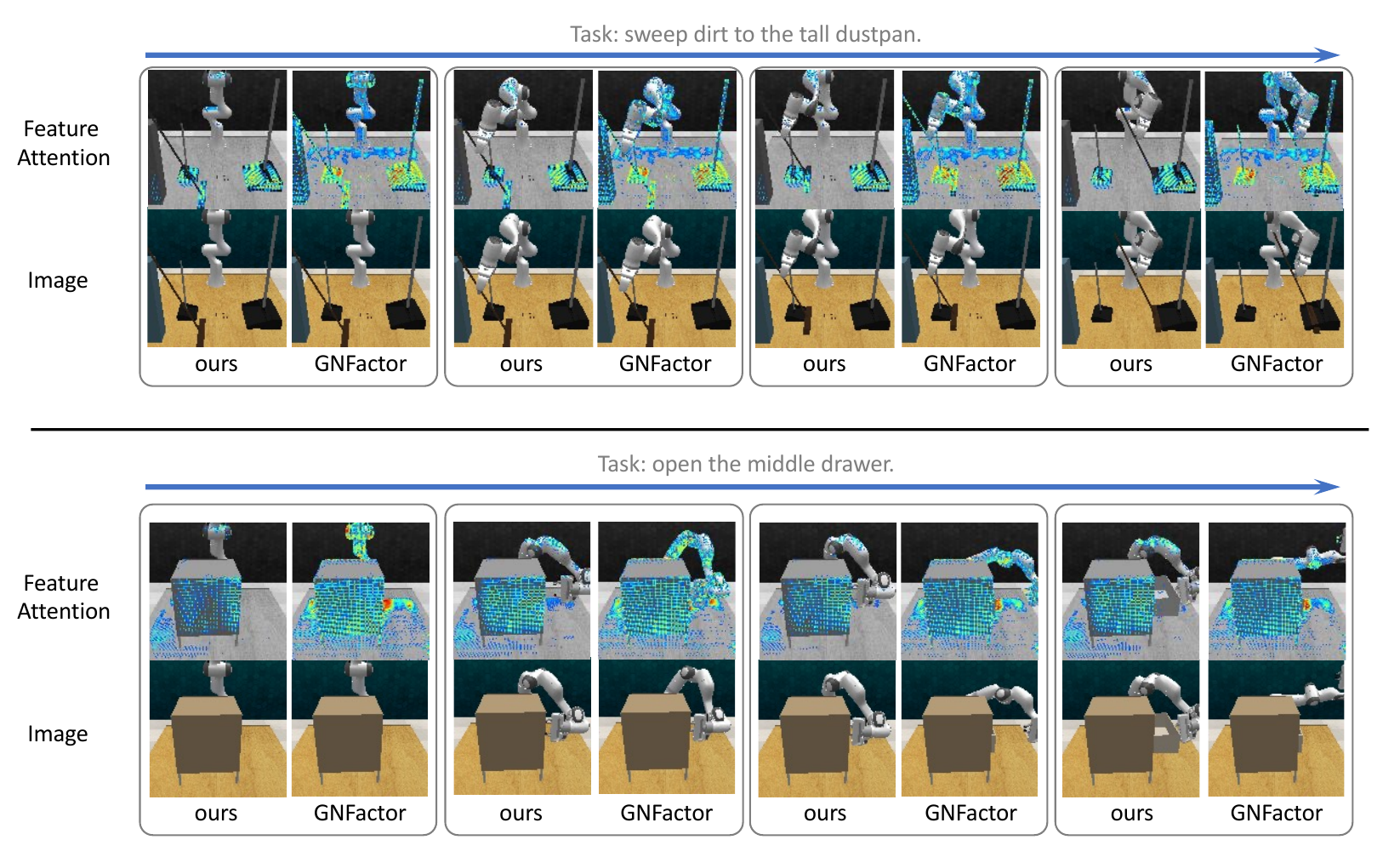}
   \caption{\textbf{Visulization Comparison of GNFactor~\cite{Ze2023GNFactor} and Ours.} }
   \label{fig:gnf_vis}
\end{figure*}

\subsection{Further Experiment Analysis}

\begin{table*}[t!]
    \centering
    \begin{adjustbox}{width=0.99\textwidth}
    \begin{tabular}{cccccccccccc}
        \toprule
        & Avg.   & close  & open  & sweep to  & meat off  & turn  & slide  & put in\  & drag  & push  & stack \\
        & Length  & jar  & drawer  & dustpan  & grill  & tap  & block  & drawer  & stick  & buttons  & blocks \\
        \midrule
         
        GNFactor & 17.0 & 19.4 & 9.9 & 16.0 & 15.1 & \textbf{11.0} & 21.1 & \textbf{16.8} & 12.8 & 23.8 & 23.8\\
        \textbf{GNFactor w/ Adapter} & \textbf{15.9} & \textbf{19.3} & \textbf{6.7} & \textbf{12.1} & \textbf{14.4} & 11.7 & \textbf{19.4} & 20.7 & \textbf{9.1} & \textbf{21.6} & \textbf{23.6} \\
        \bottomrule
    \end{tabular}
    \end{adjustbox}
    \caption{\textbf{Episode Length}. 
    We report the average episode length 
    across 5 evaluation seeds. 
    As observed, by focusing on the more relevant portion 
    of the scene with the task-aware representations, 
    action efficiency is also improved.}
    \label{tab:keypose}
\end{table*}

\paragraph{Experiments on Episode Length.}
While analyzing the success rate, we also collected statistics on the episode length of the evaluation episodes. The episode length refers to the average number of predicted keyposes or steps in each episode. A shorter episode length indicates fewer steps needed to complete a task, suggesting a more efficient prediction method. As shown in Tab.~\ref{tab:keypose}, 
our method achieves a shorter episode length than GNFactor in most tasks, on average, 6.4\% fewer steps across 10 tasks, which implies that the policy network makes more precise predictions and thus performs more efficiently.

\paragraph{Analysis of Reproducing Results of 3D Diffuser Actor.}
Due to differences in input image resolution, our reproduced results are slightly inferior to those proposed by 3D Diffuser Actor~\cite{ke20243d}. From the experimental results, it can be observed that for tasks where fine geometric details are crucial, such as "close jar," "meat off grill," and "stack blocks," our reproduced results perform poorly. This is consistent with the lack of detailed information in our data. On the other hand, we find that for tasks that only require determining the general position of an object, such as "put in drawer" and "push buttons", our reproduced results significantly outperform those reported in the original paper. This highlights the significant influence that different representations (determined by input data) have on the current process of robot learning.

\paragraph{Ablation Results Analysis.}
From the comparison of ablation results, we observe that using only the task objective as the sole condition for the hypernetwork often leads to worse performance than the original codebase. We believe this is because, in RLBench, the ten selected tasks have limited variation in task objectives, with each variation corresponding to a unique task objective. As a result, training tends to lead to the hypernetwork memorizing the task objective rather than generalizing, turning the hypernetwork into a container for memorizing a few sets of parameters instead of a tool for dynamically adjusting the information extraction process. Consequently, the entire network is prone to significant overfitting, leading to poor evaluation results.

\paragraph{Analysis on Limited Improvement Compared to 3D Diffuser Actor Codebase.}
The experiments show that compared to our significant improvement on GNFactor framework, our method has limited improvement over the 3D Diffuser Actor. Through visualization of the representation compared to the input image, we observe that, compared to the representation of the 3D Diffuser Actor, our representation is primarily focused on task-relevant areas, while the 3D Diffuser Actor's representation is more dispersed. From this perspective, our representation should significantly outperform that of the 3D Diffuser Actor during task execution. However, the final experimental results show limited improvement. We believe this is because the diffusion policy network has a strong capability for information extraction. During training, the diffusion policy not only extracts task-relevant information from the pre-trained backbone features but also further predicts action outcomes based on this information. Therefore, although our representation is better suited for learning manipulation tasks, the powerful policy network largely bridges the gap. In contrast, when using a relatively less powerful policy network, such as the Perceiver Actor, the performance improvement brought by the HyperTASR becomes much more significant.

\paragraph{Failure Case Analysis.} In simulation experiments, the failure usually appears when accurate operation on tiny objects is needed. We conduct experiments on 128$\times$128 resolution and 256$\times$256 resolution, from which we observe that with higher resolution, the average success rate increased from 78.5\% to 81.3\%. Therefore, we believe that the capability of manipulating tiny objects are highly related to the input sensory data resolution. For real robot experiments, we define task success by finishing the task without significantly changing the position of other unrelated objects in the scene. In actual evaluation, many failures are caused by changing the position of other unrelated objects due to we do incorporate collision loss in real robot experiments.

\subsection{Additional Visualization}

\begin{figure*}[htbp]
  \centering
   \includegraphics[width=0.9\linewidth]{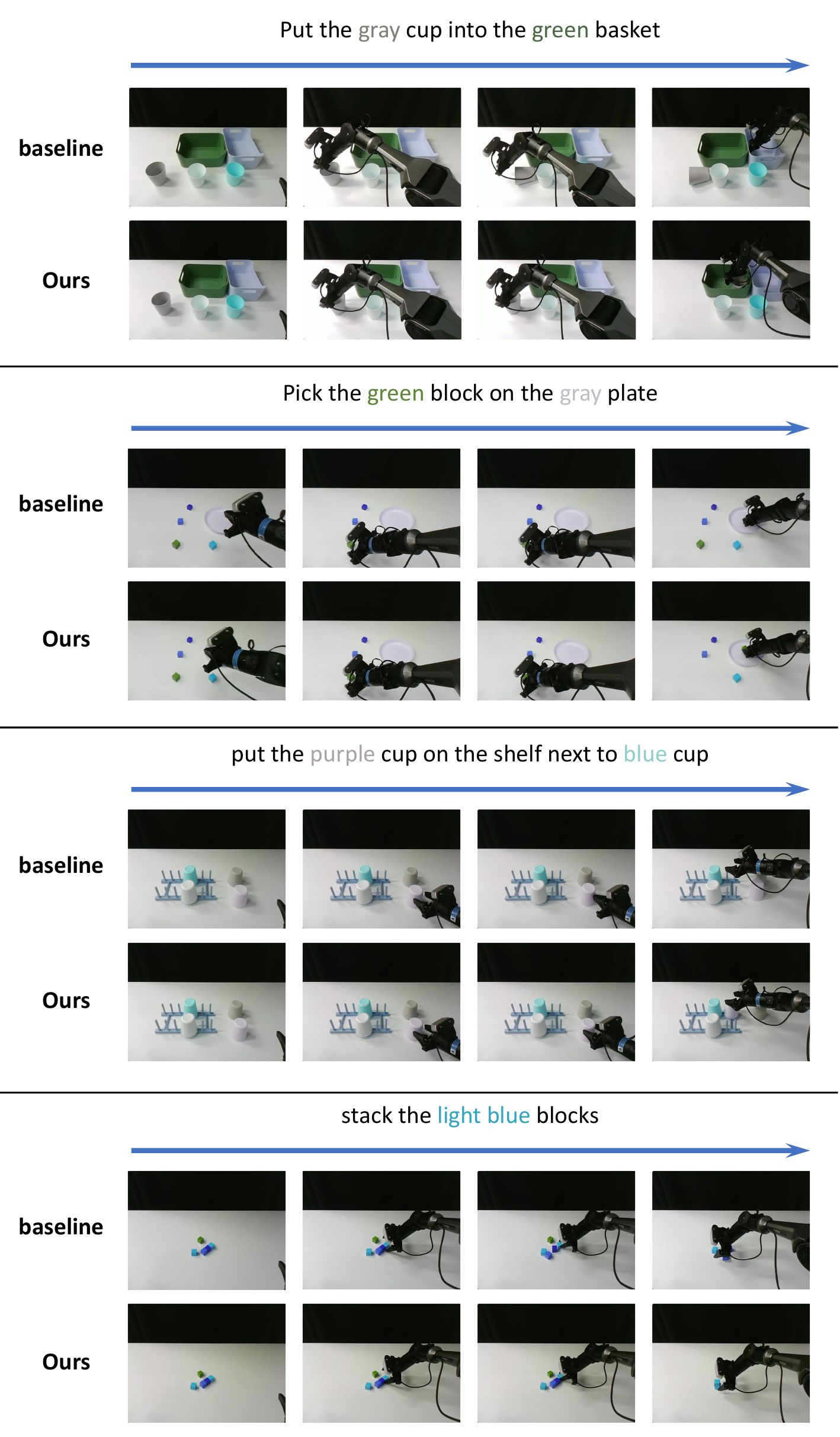}
   \caption{\textbf{Real World Task Execution Comparison of 3D Diffuser Actor and Ours.} }
   \label{fig:real_visual}
\end{figure*}

We provide additional visualization results to further demonstrate the effectiveness of our approach. First, we compare the gradient visualizations of our method and GNFactor across more tasks in Fig.~\ref{fig:gnf_vis}. From these results, it can be observed that, compared to GNFactor, our representation's attention map is more focused on task-relevant objects, whereas GNFactor's representation tends to allocate some attention to the background and objects unrelated to the task.

\begin{figure*}[htbp]
  \centering
   \includegraphics[width=0.99\linewidth]{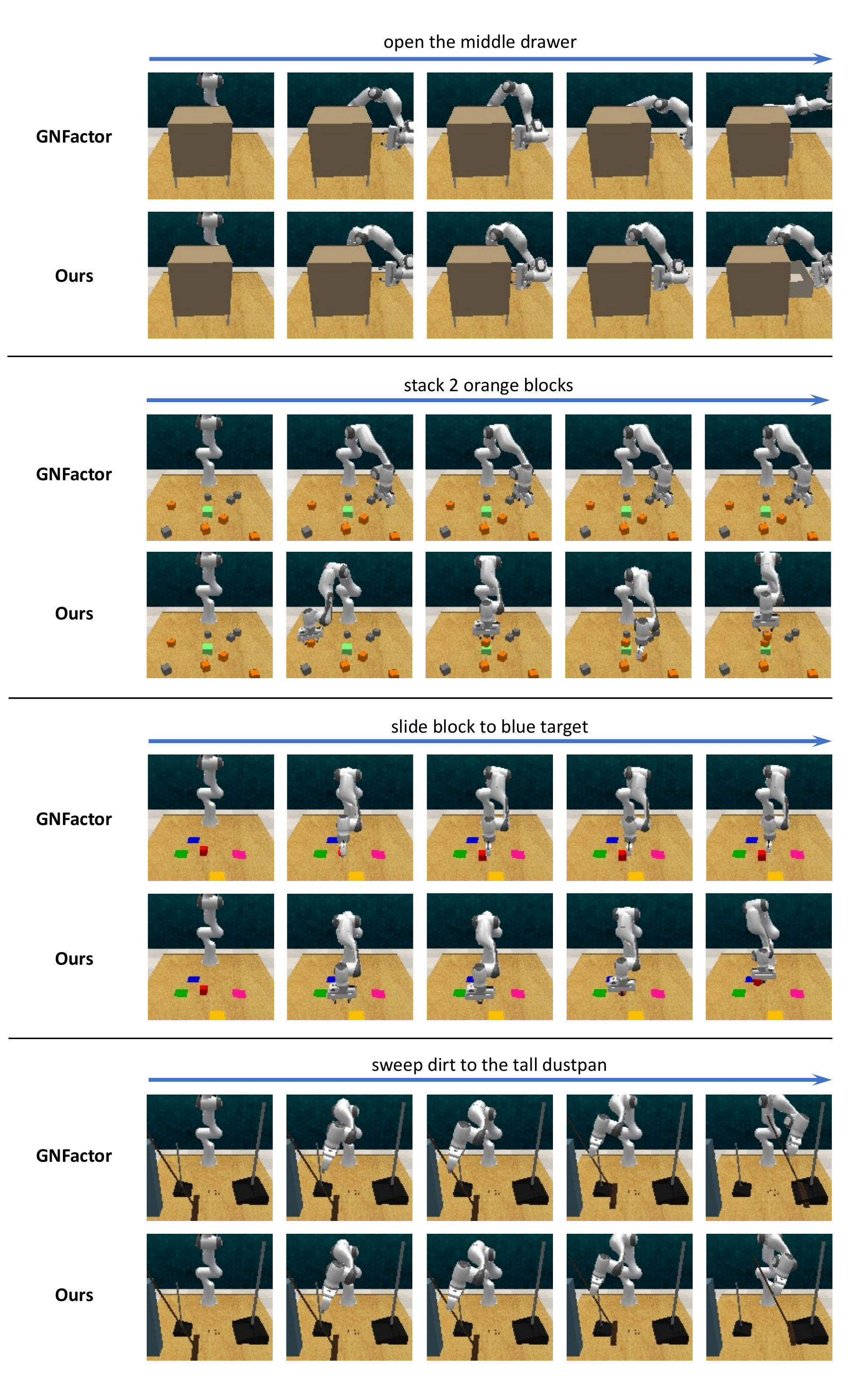}
   \caption{\textbf{Task Execution Comparison of GNFactor~\cite{Ze2023GNFactor} and Ours.} }
   \label{fig:gnf_rgb}
\end{figure*}

Next, we present a comparison of task execution between our method and GNFactor in Fig.~\ref{fig:gnf_rgb}. We provide RGB image sequences of the action execution. It can be seen that, compared to GNFactor, our approach more accurately identifies the locations of task-relevant objects, enabling more precise action execution and ultimately leading to successful task completion. In contrast, GNFactor often fails to complete the task due to getting stuck after an incorrect action execution.
Meanwhile, we present a comparison of the real-world task execution of 3D Diffuser Actor and our HyperTASR in Fig.~\ref{fig:real_visual}. More comparison are shown in Supplementary Video.

\begin{figure*}[htbp]
  \centering
   \includegraphics[width=\linewidth]{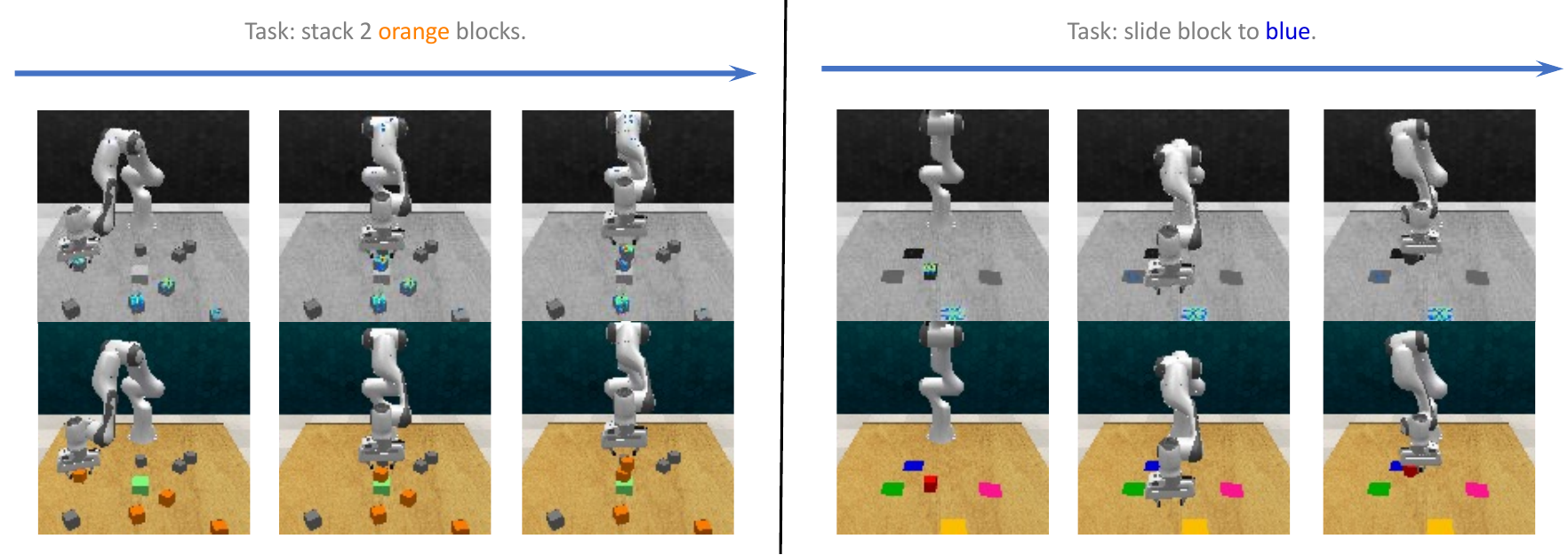}
   \caption{\textbf{Visualization Comparison regarding Task Progress for GNFactor~\cite{Ze2023GNFactor} with our HyperTASR.} }
   \label{fig:gnf_progress}
\end{figure*}

We also present the change in our representation during the action execution process in GNFactor tasks in Fig.~\ref{fig:gnf_progress}. Specifically, for the "stack blocks" task, our method shows high attention on a target block before placing it, and once the block is successfully placed, the attention on it significantly decreases. This indicates that the information regarding the block becomes less important after its placement in the context of completing the task.

\begin{figure*}[htbp]
  \centering
   \includegraphics[width=\linewidth]{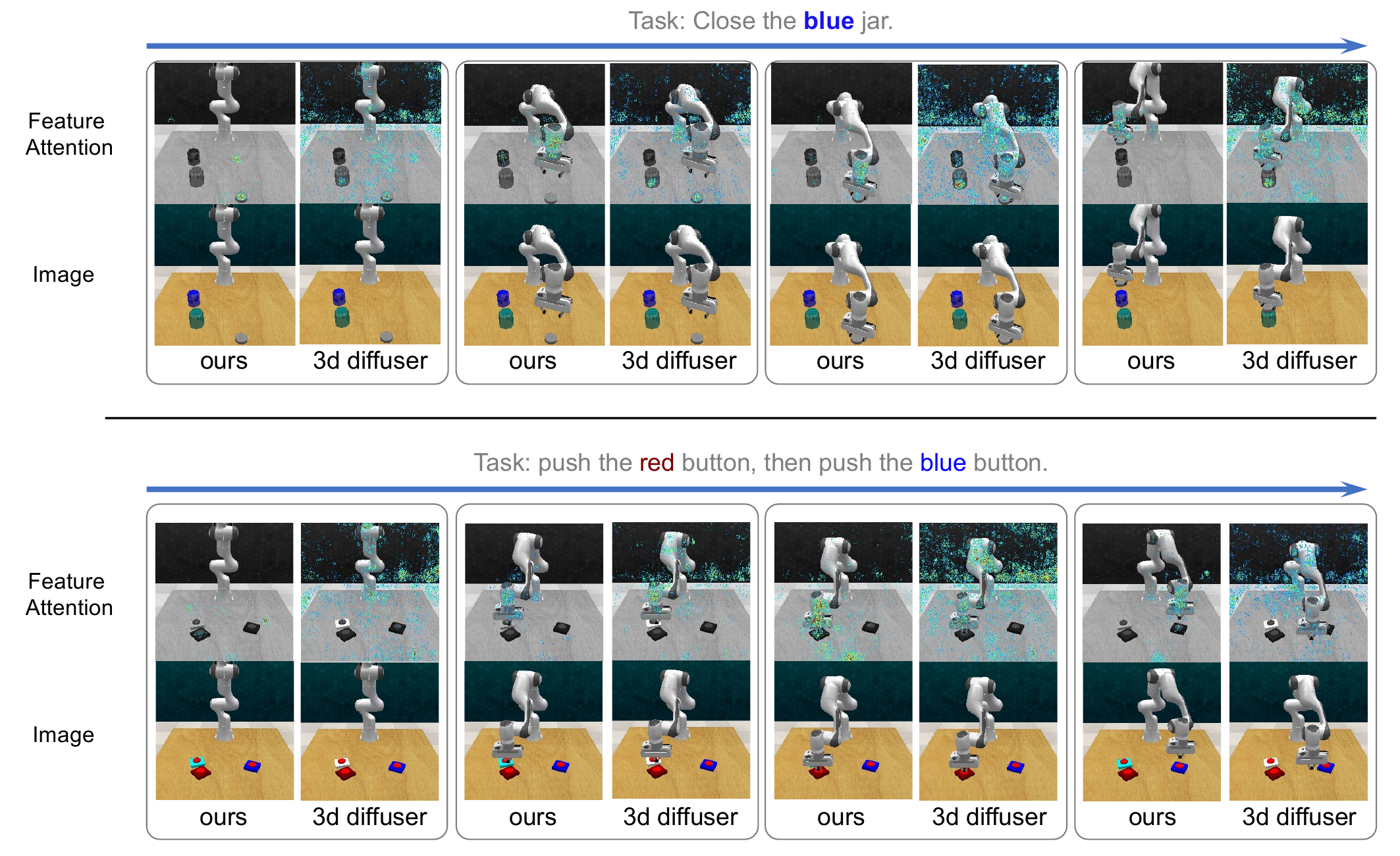}
   \caption{\textbf{Visualization Comparison of 3D Diffuser Actor~\cite{ke20243d} and Ours.} }
   \label{fig:3d_vis}
\end{figure*}

Meanwhile, we provide visualizations of the gradients of our representation versus the input image for the 3D Diffuser Actor in Fig.~\ref{fig:3d_vis}. It is evident that, compared to the 3D Diffuser Actor, our method's attention is much more concentrated.

\begin{figure*}[htbp]
  \centering
   \includegraphics[width=\linewidth]{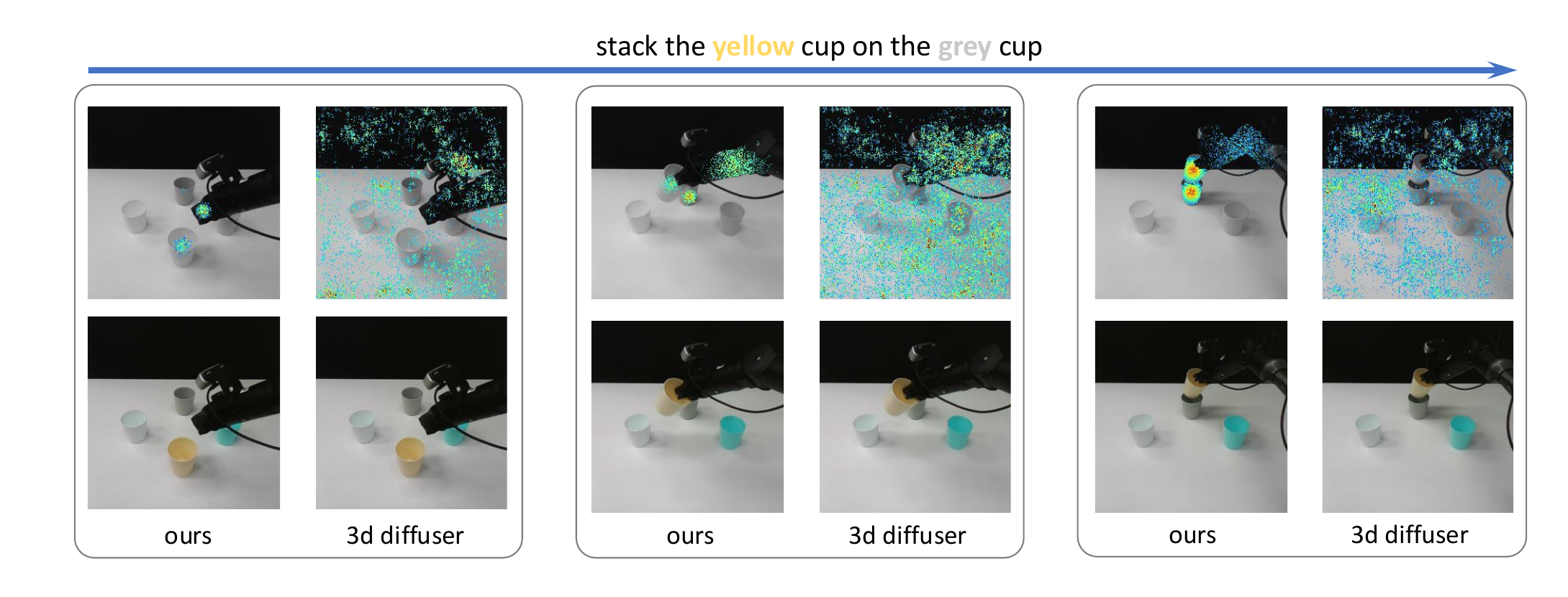}
   \caption{\textbf{Real Robot Visualization Comparison of 3D Diffuser Actor~\cite{ke20243d} and Ours.} }
   \label{fig:real_vis}
\end{figure*}

In addition, we provide attention visualization of real world experiments in Fig.~\ref{fig:real_vis}. We compute the gradient of the representation with respect to the input image. We can observe that, compared with the 3D Diffuser Actor with the attention spread through the entire image, HyperTASR is much more focused on task-related objects. Meanwhile, during the task execution, we can observe that initially, attention is focused on the yellow cup and the gripper. As the yellow cup has been picked, the attention switches to the grey cup and the robotic arm. Finally, in the stacking process, the representation focuses on two cups again. This proves our HyperTASR generates representations that dynamically adapt as the task progresses.
In Fig.~\ref{fig:real_vis}, we visualize attention in real robot experiments by computing the gradient of the learned representation with respect to the input image of the training set. Unlike the 3D Diffuser Actor, whose attention is diffusely distributed across the scene, HyperTASR concentrates its attention on task-relevant objects. During the early grasping phase, attention is tightly focused on the yellow cup and the gripper. Once the yellow cup is lifted, attention shifts to the grey cup and the robotic arm. Finally, as the stacking motion commences, the model’s attention returns to both cups. These observations demonstrate that HyperTASR produces dynamic, task-aware representations that track the evolving focus requirements throughout task execution.

\end{document}